\documentclass[runningheads,a4paper]{llncs}

\usepackage{amssymb}
\usepackage{graphicx}
\usepackage{booktabs}

\usepackage{url}
\urldef{\mailsa}\path|ys.zhang@connect.ust.hk |
\urldef{\mailsb}\path|achung@cse.ust.hk|

\begin{document}

\mainmatter

\title{Deep supervision with additional labels for retinal vessel segmentation task}

\titlerunning{Deep supervision with additional labels for retinal vessel segmentation task}

\author{Yishuo ZHANG and Albert C.S. Chung}

\authorrunning{Yishuo ZHANG and Albert C.S. Chung}

\institute{Lo Kwee-Seong Medical Image Analysis Laboratory,\\
Department of Computer Science and Engineering,\\
The Hong Kong University of Science and Technology, Hong Kong\\
\mailsa, \mailsb}

\toctitle{Deep supervision with additional labels for retinal vessel segmentation task}
\tocauthor{}
\maketitle

\begin{abstract}
Automatic analysis of retinal fundus images is of vital importance in diagnosis tasks of retinopathy. Segmenting vessels accurately is a fundamental step in analysing retinal images. However, it is usually difficult due to various imaging conditions, low image contrast and the appearance of pathologies such as micro-aneurysms. In this paper, we propose a novel method with deep neural networks to solve this problem. We utilize U-net with residual connection to detect vessels. To achieve better accuracy, we introduce an edge-aware mechanism, in which we convert the original task into a multi-class task by adding additional labels on boundary areas. In this way, the network will pay more attention to the boundary areas of vessels and achieve a better performance, especially in tiny vessels detecting. Besides, side output layers are applied in order to give deep supervision and therefore help convergence. We train and evaluate our model on three databases: DRIVE, STARE, and CHASEDB1. Experimental results show that our method has a comparable performance with AUC of 97.99\% on DRIVE and an efficient running time compared to the state-of-the-art methods.
\end{abstract}

\section{Introduction}

Retinal vessels are commonly analysed in the diagnosis and treatment of various ophthalmological diseases. For example, retinal vascular structures are correlated to the severity of diabetic retinopathy \cite{jelinek2009automated}, which is a  cause of blindness globally. Thus, the precise segmentation of retinal vessels is of vital importance. However, this task is often extremely challenging due to the following factors \cite{fraz2012approach}: 1. The shape and width of vessel vary, which cannot be represented by a simple pattern; 2. The resolution, contrast and local intensity change among different fundus images, increasing the difficulty of segmenting; 3. Other structures, like optical disks and lesions, can be interference factors; 4. Extremely thin vessels are hard to detect due to the low contrast and noise.

In recent years, a variety of methods have been proposed to solve retinal vessel segmentation tasks, including unsupervised methods \cite{marin2011new} and supervised methods \cite{orlando2014learning}. Although promising performances have been shown, there is still some room for improvement. As we mentioned before, tiny capillaries are hard to find and missing these can lead to low sensitivity. Besides, methods that need less running time are preferred in clinical practice. In this paper, we aim to design a more effective and efficient method to tackle these problems.

The emergence of deep learning methods provides a powerful tool for computer vision tasks and these kinds of methods have outperformed other methods in many areas. By stacking convolutional layers and pooling layers, networks can gain the capacity to learn the very complicated representation of features. U-net, proposed in \cite{ronneberger2015u}, can deal with image patches in an end-to-end manner and therefore is widely used in medical image segmentation.

\begin{figure}[h]
	\centering
	\vspace{-10pt}
	\includegraphics[height=3cm]{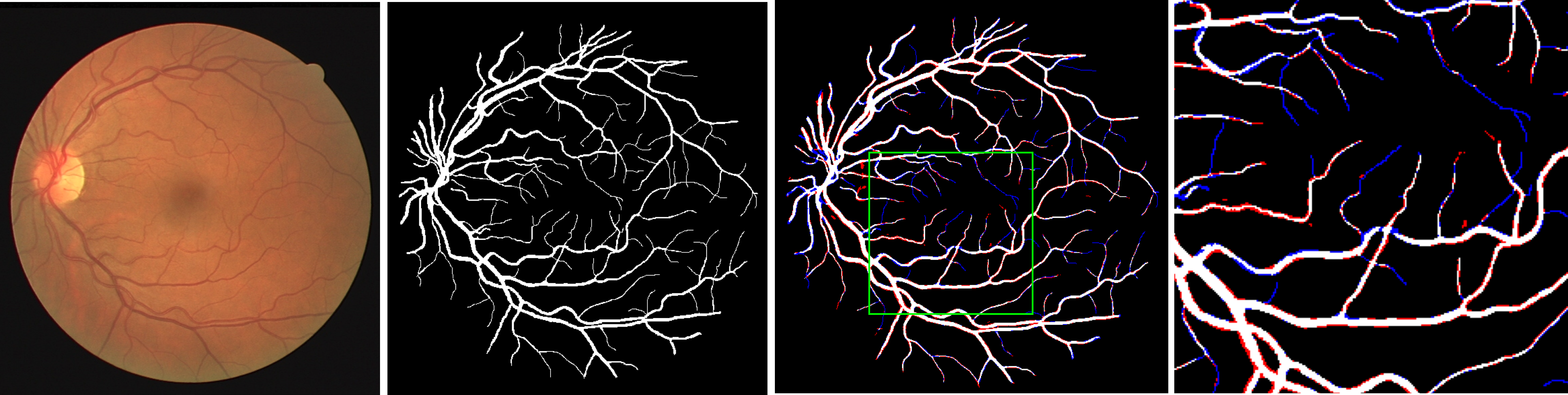}
	\caption{Images sampled from datasets. From left to right: original fundus image, ground truth, output of a single trained U-net and zoomed segment inside the green rectangle in the third image. In the last image, blue regions denote false negative, while red regions denote false positive.}
	\vspace{-10pt}
	\label{fig:exp}
\end{figure}

We analyse the output of a single trained U-net model as shown in Fig.~\ref{fig:exp}. Most mislabelled pixels come from boundaries between foreground and background. Regarding thick vessels, the background areas around vessels are easy to be labelled as positive. However regarding very thin vessels, many of these are ignored by networks and labelled as background. To tackle this problem, we process the ground truth, by labelling the boundary, thick vessels and thin vessels as different classes, which forces the networks to pay different extra attention to error-prone regions. This operation makes the original task become a harder task. If the new task can be solved by our method perfectly, then so could the original task. Besides, we also utilize deep supervision to help networks converge. 

Our main contributions are as follows:
\vspace{-5pt}
\begin{enumerate}
	\item Introducing a deep supervision mechanism into U-net, which helps networks learn a better semantically representation;
	
	\item Separating thin vessels from thick vessels during the progress of training;
	
	\item Applying an edge-aware mechanism by labelling the boundary region to extra classes, making the network focus on the boundaries of vessels and therefore get finer edges in the segmentation result.
\end{enumerate}

\section{Proposed Method}

\subsection{U-net}

\begin{figure}[h]
	\centering
	\vspace{-15pt}
	\includegraphics[height=6cm]{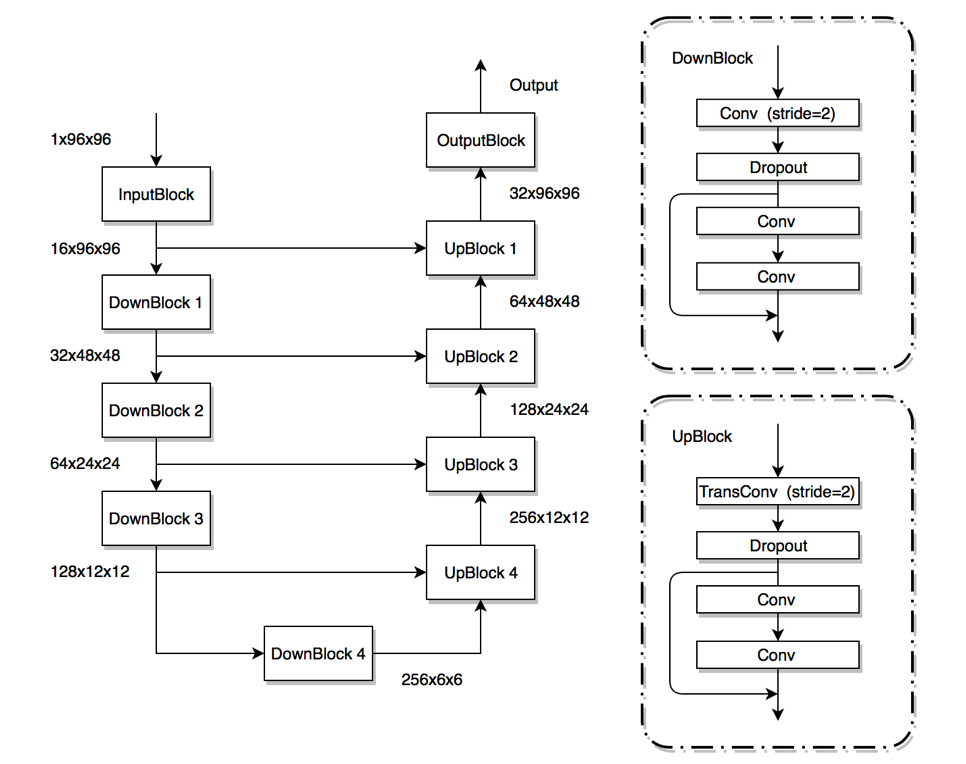}
	\vspace{-5pt}
	\caption{Architecture of a simple U-net. We annotate shapes about feature maps of each block in the format of `Channels, Width, Height'. Inner structures of DownBlock and UpBlock are shown on the right, where each Conv layer is followed by two unseen layers: a BatchNorm layer and a ReLU layer.}
	\vspace{-15pt}
	\label{fig:unet}
\end{figure}

The architecture of U-net is illustrated in Fig.~\ref{fig:unet}. The left-hand part consists of four blocks, each of which contains stacked convolutional layers (Conv) to learn hierarchical features. The size of the input feature maps is halved after each stage, implemented by a Conv layer with a stride of 2. In contrast, the number of feature channels increases when the depth increases, in order to learn more complicated representations. The right-hand part has a similar structure to the left part. The size of input feature maps is doubled after each stage by a deconvolution layer to reconstruct spatial information.

To utilize feature learned by earlier layers at subsequent layers, feature maps from the left-hand blocks will be fed into the corresponding right-hand blocks. In this way, networks can gain detailed information which may be lost in former downsampling operations but useful for fine boundary prediction. To improve the robustness and help convergence, we apply a residual connection \cite{he2016deep} inside each block, which adds feature maps before Conv layers to the output maps pixel-wisely. We also leverage Dropout and BatchNorm inside each block to reduce overfitting and gradient vanishing respectively. 

\subsection{Additional Label}

Additional labels are added to the original ground truth before training, which converts this task into a multi-class segmentation task. Firstly, we distinguish thick vessels from thin vessels (with a width of 1 or 2 pixels), implemented by an opening operation. Then, we locate the pixels near to the vessel by a dilation operation and label them to the additional class. Therefore, we have 5 classes, which are 0 (other background pixels), 1 (background near thick vessels), 2 (background near thin vessels), 3 (thick vessel) and 4 (thin vessel).

\begin{figure}[h]
	\centering
	\vspace{-10pt}
	\includegraphics[height=3cm]{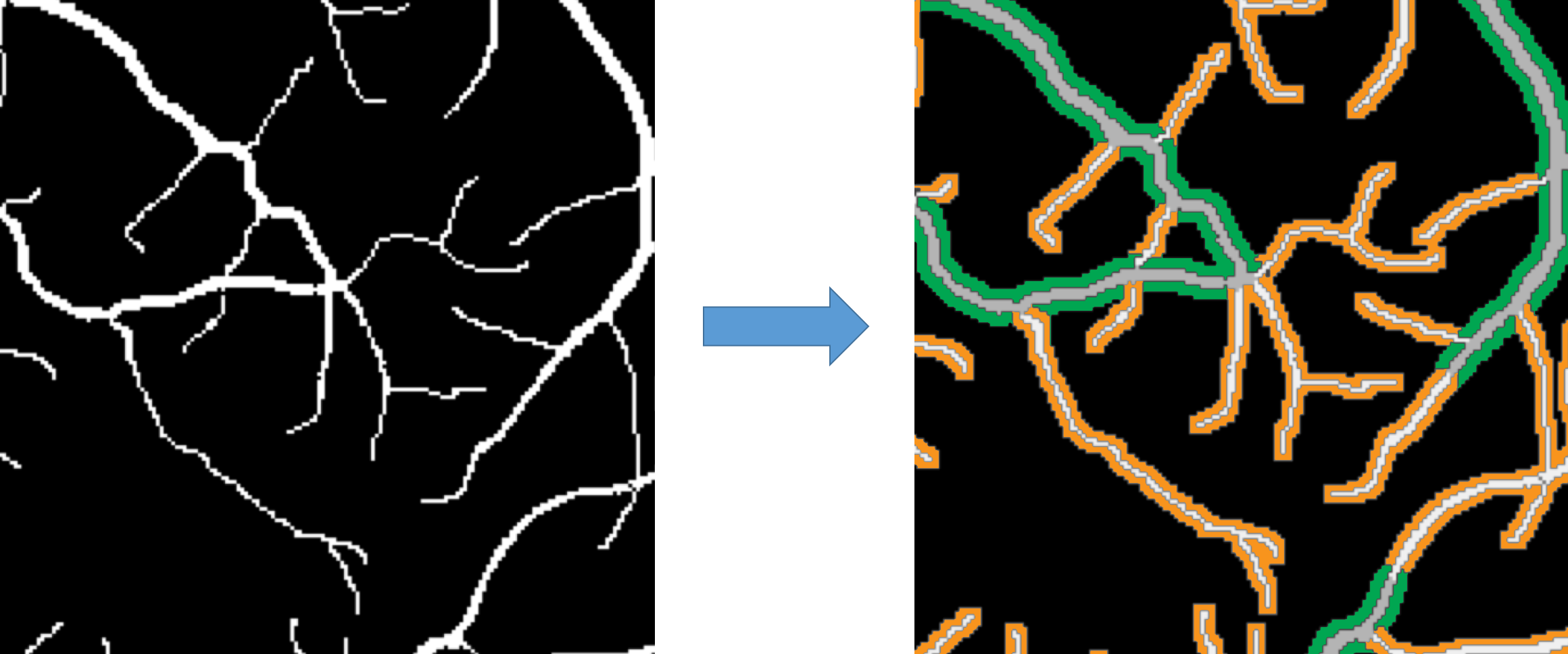}
	\caption{Generated new multi-class ground truth, where different classes are shown in different color: 0(black), 1(green), 2(orange), 3(grey), 4(white). }
	\vspace{-20pt}
	\label{fig:newlabel}
\end{figure}

The objective of this is to force the networks to treat background pixels differently. As we reported above, the boundary region is easy to be mislabelled. We separate these classes so that we can give more supervision in crucial areas by modifying the class weight in the loss function but not influencing others. Boundary classes have heavier weights in the loss function, which means that these classes will attract a higher penalty if labelled wrongly.  (Fig.~\ref{fig:newlabel})

\subsection{Deep Supervision}

\begin{figure}[h]
	\centering
	\vspace{-15pt}
	\includegraphics[height=6cm]{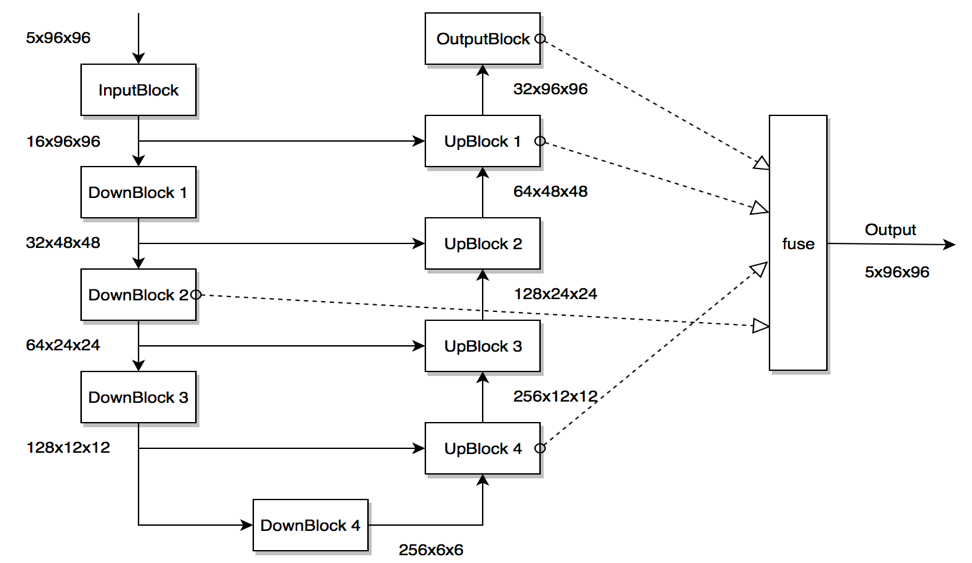}
	\caption{Diagram of our proposed method. }
	\vspace{-15pt}
	\label{fig:deeps}
\end{figure}

Deep supervision \cite{lee2015deeply} is employed to solve the problem of information loss during forward propagation and improve detailed accuracy. This mechanism is beneficial because it gives semantic representations to the intermediate layers. We implement it by adding four side output layers as shown in Fig.~\ref{fig:deeps}. The output of each side layer is compared with the ground truth to calculate auxiliary losses. Final prediction maps are generated by fusing the outputs of all four side layers.

We employ cross-entropy as loss function and calculate it for the final output as well as each side output. Owing to the amounts of different classes being imbalanced, we add a class-balanced weight for each class to correct imbalances. As we discussed before, pixels of boundaries around thick vessels and pixels of thin vessels should be given relatively heavier weights.

\begin{equation}
	\vspace{-10pt}
	CE(pred,target) = - \sum_{i}^{}weight_{i}\times target_{i}\times log(pred_{i}) \; .
	\vspace{-0pt}
\end{equation}

The total loss is defined as below, comprising of loss of fused output, losses of side outputs and L-2 regular term.

\begin{equation}
	\vspace{-10pt}
	Loss = CE(fuse,GT) + \sum_{3}^{n}CE(side_{i},GT) + \frac{\lambda}{2} \|w\|^{2} \;.
	\vspace{-5pt}
\end{equation}

\section{Experiments}

We implement our model with PyTorch library. Stochastic gradient descent algorithm (SGD) with momentum is utilized to optimize our model. The learning rate is set to 0.01 initially and halved every 100 epochs. We train the whole model for 200 epochs on a single NVIDIA GPU (GeForce Titan X). The training progress takes nearly 10 hours.

\subsection{Datasets}

We evaluate our method on three public datasets: DRIVE \cite{staal2004ridge}, STARE \cite{hoover2000locating} and CHASEDB1 \cite{fraz2012ensemble}, each of which contains images with two labelled masks annotated by different experts. We take the first labelled mask as ground truth for training and testing. The second labelled masks are used for comparison between our model and a human observer. The DRIVE dataset contains 20 training images and 20 testing images; thus, we take them as the training set and testing set respectively. STARE and CHASEDB1 datasets contain 20 and 28 images, respectively. As these two datasets are not divided for training and testing, we perform a four-fold cross-validation, following \cite{mo2017multi}.

Before feeding the original image into networks, some preprocessing operations are performed. We employ contrast-limited adaptive histogram equalization (CLAHE) to enhance the image and increase contrast. Then the whole images are cropped into patches with the size of 96*96 pixels. To augment the training data, we perform the flip, affine transformation, and noising operations randomly. In addition, the lightness and contrast of the original images are changed randomly to improve the robustness of the model.

\subsection{Results}

A vessel segmentation task can be viewed as an unbalanced pixel-wise classification task. For evaluation purpose, measurements including Specificity (Sp), Sensitivity (Se) and Accuracy (Acc) are computed. They are defined as below:

\begin{equation}
	\vspace{-5pt}
	Sp = \frac{TN}{TN+FP}\; , Se=\frac{TP}{TP+FN}\; , Acc=\frac{TP+TN}{TP+FP+TN+FN} \; ,
	\vspace{-0pt}
\end{equation}

Here TP, FN, TN, FP denote true positive, false negative, true negative and false positive, respectively. Additionally, a better metric, area under the receiver operating characteristic (ROC) curve (AUC), is used. We believe that AUC is more suitable for measuring an unbalanced situation. A perfect classifier should have an AUC value of 1.

\begin{figure}[h]
	\centering
	\vspace{-15pt}
	\includegraphics[height=3cm]{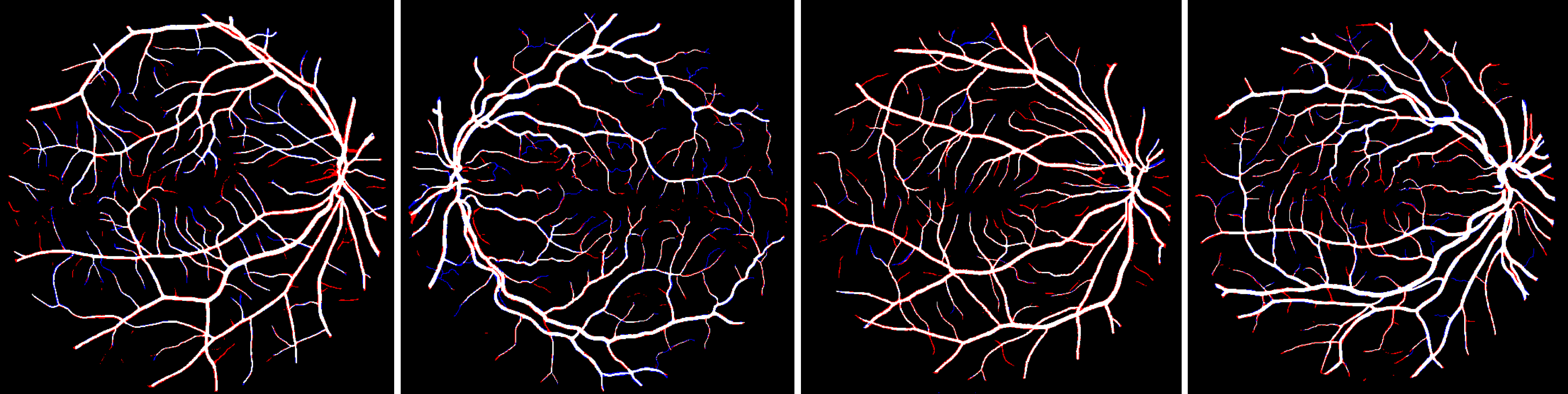}
	\vspace{-5pt}
	\caption{Examples of our experiment output. }
	\vspace{-15pt}
	\label{fig:res}
\end{figure}

\begin{figure}[h]
	\centering
	\vspace{-20pt}
	\includegraphics[height=2cm]{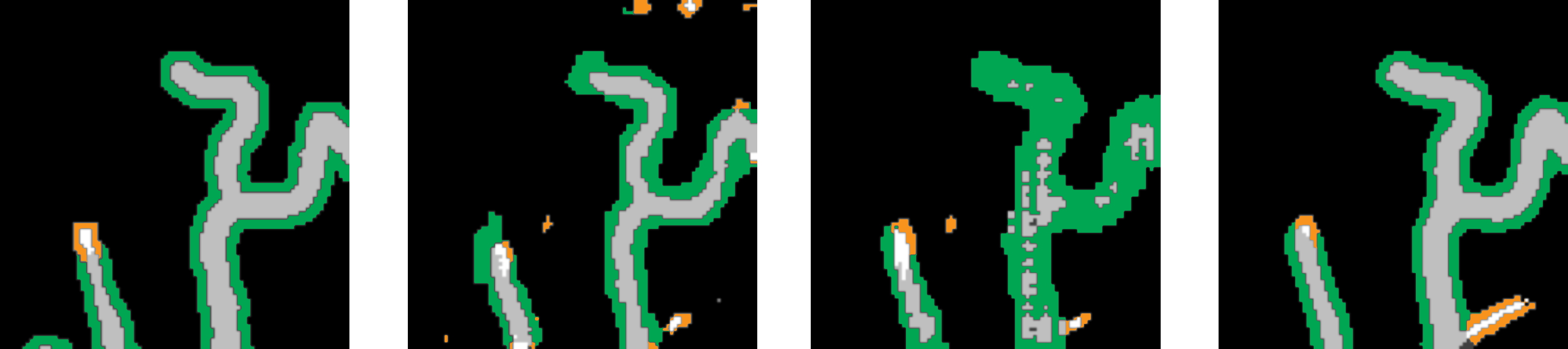}
	\caption{Comparison between side output and ground truth. From left to right: Ground Truth, 2 side outputs, and final prediction.}
	\vspace{-15pt}
	\label{fig:vis}
\end{figure}

We have three observations: 1. Even if the side output cannot locate the vessel, they can locate the boundary region, which can help to find the vessel in final output precisely as a guide. 2. As the resolution of the side output is lower, the tiny vessel may be missed but the boundary region is more distinct and easier to find. This shows mutual promotion between the additional label and deep supervision. 3. The boundary of the boundary region is not refined, but it does not affect the prediction because we will take all of the boundary regions as background. (Fig.~\ref{fig:res}, Fig.~\ref{fig:vis})

\begin{table}[h]
	\centering
	\vspace{-10pt}
	\caption{Performance comparison with simple U-net on dataset DRIVE}
	\begin{tabular}{cccc}
		\toprule
		Methods&AUC of all vessels&AUC of thick vessels&AUC of thin vessels\\
		\midrule	
		Simple U-net&0.9736&0.9830&0.8678\\
		Our method&0.9799&0.9897&0.9589\\
		\bottomrule
	\end{tabular}
	\vspace{-10pt}
	\label{table1}
\end{table}

To validate the effect of our idea, we perform comparison experiments with a simple U-net. With additional label and well-designed deep supervision, our method has better capabilities of detecting vessels, especially for capillaries. AUC of thin vessels has been increased by 9.11\%, as shown in Table~\ref{table1}.

\subsection{Comparison}

We report performances of our method in respect to the aforementioned metrics, compared with other state-of-the-art methods, as shown in Table~\ref{table2} and Table~\ref{table3}.
\begin{table}[h]
	\centering
	\vspace{-10pt}
	\caption{Performance comparison on the DRIVE dataset}
	\vspace{-5pt}
	\begin{tabular}{c|c|c|c|c}
		\toprule
		Methods&Acc&Sp&Se&AUC\\
		\midrule	
		2nd Observer&0.9472&0.9730&0.7760&N.A.\\
		Fraz et al. \cite{fraz2012ensemble}	&0.9480	&\textbf{0.9807}	&0.7406	&0.9747\\
		Liskowski et al. \cite{liskowski2016segmenting} &\textbf{0.9535}	&\textbf{0.9807}	&0.7811	&0.9790\\
		Mo et al. \cite{mo2017multi}	&0.9521	&0.9780	&0.7779	&0.9782\\
		Leopold et al. \cite{leopold2017pixelbnn}	&0.9106	&0.9573	&0.6963	&0.8268\\
		Our method	&0.9504	&0.9618	&\textbf{0.8723}	&\textbf{0.9799}\\
		\bottomrule
	\end{tabular}
	\vspace{0pt}
	\label{table2}
\end{table}
\begin{table}[h]
	\centering
	\vspace{-5pt}
	\caption{Performance comparison on STARE and CHASEDB1 datasets}
	\vspace{-5pt}
	\begin{tabular}{c|c|c|c|c|c|c|c|c}
		\toprule
		&\multicolumn{4}{|c|}{STARE}&\multicolumn{4}{|c}{CHASEDB1}\\
		Methods&Acc&Sp&Se&AUC&Acc&Sp&Se&AUC\\
		\midrule	
		2nd Observer&0.9353&0.9387&\textbf{0.8951}&N.A.&0.9560&0.9793&0.7425&N.A.\\
		Fraz et al. \cite{fraz2012ensemble}	&0.9534	&0.9763	&0.7548	&0.9768&0.9468	&0.9711	&0.7224	&0.9712\\
		Liskowski et al. \cite{liskowski2016segmenting} &\textbf{0.9729}	&0.9862	&0.8554	&\textbf{0.9928}&0.9628	&0.9836	&0.7816	&0.9823\\
		Mo et al. \cite{mo2017multi}	&0.9674	&0.9844	&0.8147	&0.9885&0.9599	&0.9816	&0.7661	&0.9812\\
		Leopold et al. \cite{leopold2017pixelbnn}	&0.9045	&0.9472	&0.6433	&0.7952&0.8936	&0.8961	&\textbf{0.8618}	&0.8790\\
		Our method	&0.9712	&\textbf{0.9901}	&0.7673	&0.9882&\textbf{0.9770}	&\textbf{0.9909}	&0.7670	&\textbf{0.9900}\\
		\bottomrule
	\end{tabular}
	\vspace{-15pt}
	\label{table3}
\end{table}
We have highlighted the highest scores for each column. Our method achieves the highest Sensitivity on the DRIVE dataset and the highest Specificity on the other two datasets. Due to the differences of inherent errors among datasets and the class imbalance, we prefer using AUC as an equatable metric for comparison. Our method has the best performance on the DRIVE and CHASEDB1 datasets in terms of AUC.

\begin{table}[h]
	\centering
	\vspace{-15pt}
	\caption{Time comparison with other methods}
	\vspace{-5pt}
	\begin{tabular}{ccc}
		\toprule
		Method	&Training Time (h)	&Running Time (s)\\
		\midrule	
		Liskowski et al.~\cite{liskowski2016segmenting}	&8	&92\\
		Mo et al.~\cite{mo2017multi}	&10	&0.4\\
		Our method	&10	&0.9\\
		\bottomrule
	\end{tabular}
	\vspace{-20pt}
	\label{table4}
\end{table}

In terms of running time, our method is also computationally efficient when compared to other methods. (Table.~\ref{table4}) Our proposed method can deal with an image size of 584*565 in 1.2s, much faster than the method proposed in \cite{liskowski2016segmenting}. This benefit is obtained from our method by using the U-net architecture which works from patch to patch, instead of using a patch to predict the central pixel alone. The method proposed in \cite{mo2017multi} is a little faster than ours, as their network has less up-sampling layers. However, removing up-sampling leads to a decrease in fine prediction and especially sensitivity. By overall consideration, we choose proper numbers of layers as used in our presented method, which can achieve the best performance with highly acceptable running time.

\section{Conclusion}

In this paper, we propose a novel deep neural network to segment retinal vessel. To give more importance to boundary pixels, we label thick vessels, thin vessels and boundaries into different classes, which makes a multi-class segmentation task. We use a U-net with residual connections to perform the segmentation task. Deep supervision is introduced to help the network learn better features and semantic information. Our method offers a good performance and efficient running time compared to other state-of-the-art methods, which can give high efficacy in clinical applications.

\bibliography{ref}
\bibliographystyle{splncs03}

\end{document}